\documentclass[12pt]{article}
\usepackage{epsfig}
\usepackage{amsmath,amssymb}
\usepackage{amsthm}
\usepackage{setspace}
\usepackage{amsbsy}
\usepackage{natbib}
\usepackage{booktabs}
\newcommand{\ra}[1]{\renewcommand{\arraystretch}{#1}}

\newcommand{\T}{^{\rm T}}

\begin{document}
\title{Quantile universal threshold: model selection at the detection edge for high-dimensional linear regression}


\author{Jairo Diaz Rodriguez \footnote{2-4 rue du Li\`evre, CP 64, 1211 Gen\`eve 4, Switzerland; jairo.diaz@unige.ch}\\{\em Department of Mathematics, University of Geneva} \\
and
\\
Sylvain Sardy \footnote{2-4 rue du Li\`evre, CP 64, 1211 Gen\`eve 4, Switzerland; sylvain.sardy@unige.ch}\\{\em Department of Mathematics, University of Geneva}}
\date{}

 \maketitle

\begin{quote}
To estimate a sparse linear model from data with Gaussian noise, 
consilience from lasso and compressed sensing literatures is that thresholding estimators like lasso and the Dantzig selector have
the ability in some situations to identify with high probability part of the significant
covariates asymptotically, and are numerically tractable thanks to convexity.

Yet, the selection of a threshold parameter $\lambda$ remains crucial in practice. To that aim we propose Quantile Universal Thresholding,
a selection of $\lambda$ at the detection edge. 
We show with extensive simulations and real data that an excellent compromise
between high true positive rate and low false discovery rate is achieved, leading also to good predictive risk.

\end{quote}




\newpage

\section{Introduction}
\label{sct:intro}

The goal of regression is to relate covariates ${\bf x}=(x_{1}, \ldots, x_{P})$ to an associated response $y$ through a rule $\hat \mu({\bf x})$ estimated
based on a set of training measurements $\{(y_n, {\bf x}_n)\}_{n=1,\ldots,N}$.
In Chemometrics, the covariates are spectrometer measurements and the response can be the octane level of a fuel sample: predicting the octane level from
its cheap spectrometer measurements can save time and money in comparison with tedious and time consuming mechanical techniques.
In Genetics, the expression levels are measured on a large number of genes and gathered in a microarray:
identifying and testing which genes are significantly related to a response of interest can reveal important to cure certain genetic diseases.
A common pattern between modern recording devices is that the number of covariates $P$ measured per sample is large, and is getting larger. It is therefore getting common
to analyze data where $P$ exceeds the sample size $N$, which makes the task of Statistics more difficult than when $N>P$.

We concentrate on the most commonly used  model when the rule relating ${\bf x}_n$ to $y_n$ is assumed to be a linear combination of the covariates
with unknown coefficients~${\boldsymbol \beta}^0$,
\begin{equation} \label{eq:linearmodel}
y_n = \mu_n^0 + \epsilon_n \quad {\rm with} \quad \mu_n^0 = \sum_{p=1}^P x_{np} \beta_p^0={\bf x}_n^{\rm T} {\boldsymbol \beta}^0, \quad n=1,\ldots,N,
\end{equation}
where $\epsilon_n$ are the model errors often assumed i.i.d.~Gaussian ${\rm N}(0,\sigma^2)$.
In matrix notation, we assume ${\bf y}=X {\boldsymbol \beta}^0+{\boldsymbol \epsilon}$, where $X$ is an $N \times P$ matrix. 
Estimation of the coefficients ${\boldsymbol \beta}^0$ by least squares fails when $P>N$, not only for the lack of a unique estimate, but also
for bad bias-variance trade-off and poor predictive performance.

To overcome the drawbacks of least squares estimation motivated by the seminal papers of \citet{Jame:Stei:esti:1961} and \citet{Tikhonov63},
a considerable amount of literature has concentrated over the last fifty years on the estimation of the coefficients ${\boldsymbol \beta}^0$ by regularization techniques
that aim at decreasing the variance by introducing some bias for a better prediction error.
The prominent regularized estimators are ridge regression \citep{ridgeHK}, principal component regression \citep{Mass:prin:1965}, partial least squares \citep{Wold:esti:1966},
subset variable selection and, more recently, the lasso \citep{Tibs:regr:1996} and the Dantzig selector \citep{Cand:Tao:dant:2007}.
The last three estimators rely on the important assumption that the true model
\begin{equation}\label{eq:Model*}
{\cal S}^0=\{p\in\{1,\ldots,P\}:\ \beta_p^0 \neq 0\} 
\end{equation}
is \emph{sparse} with sparsity parameter $s^0=|{\cal S}^0|$.
Assuming sparsity for the underlying model makes the task of identifying a model more reasonable when $P>N$.

All these regularization techniques are governed by a parameter we will call $\lambda$, that controls the bias-variance trade-off.
It must be selected from the data.
Criteria to select $\lambda$ are mainly of two kinds, whether it is based on:
\begin{enumerate}
 \item predictive performance of  $\hat \mu^{\rm new}=({\rm x}^{\rm new})\T \hat {\boldsymbol \beta}(\lambda)$  with  new covariates ${\rm x}^{\rm new}$.
 This is the situation for the Chemometrics example discussed above:
 predicting octane level with accuracy is the goal. A measure of good prediction is the predictive risk between $\hat \mu^{\rm new}$ and $y^{\rm new}$.
 \item identify relevant covariates, that is identify (parts of) ${\cal S}^0$ in (\ref{eq:Model*}).
 This is the situation for the Genetics example discussed above:
 by setting $\hat \beta_p(\lambda)=0$ for most $p$, sparse estimation selects covariates as those
 with associated estimated coefficients different from zero.
 To measure the pertinence of variable selection techniques, many criteria can be used. Prominent ones are high true positive rate and low false discovery rate \citep{Benj:Hoch:1995} defined by
 \begin{equation} \label{eq:TPRandFDR}
   {\rm TPR}= \frac{| \hat {\cal S}_\lambda \cap {\cal S}^0|}{|{\cal S}^0|} \quad {\rm and} \quad {\rm FDR}=\frac{|\hat {\cal S}_\lambda \cap \bar {\cal S}^0|}{|\hat {\cal S}_\lambda|},
 \end{equation}
 where ${\cal S}^0$ is defined in (\ref{eq:Model*}), $\bar {\cal S}^0=\{p\in\{1,\ldots,P\}:\ \beta_p^0 = 0\} $, and $\hat {\cal S}_\lambda=\{p\in\{1,\ldots,P\}:\ \hat \beta_p(\lambda) \neq 0\} $.
 When their denominator is null TPR is equal to one and FDR is equal to zero.
\end{enumerate}
Of criteria~1 and~2, this paper is concerned with the latter, when identification of the relevant covariates is as important as prediction, if not more.
Although correlated, these criteria do not lead to the same optimal coefficients estimates. 
In other words, a selected $\lambda$ may be optimal for one criterion but not for the other.

\subsection{Key properties of the lasso}

What makes the project of achieving high TPR and low FDR realistic in a situation of low sample size $N$ and large number $P$ of covariates
is the consilience from lasso (noisy setting) \citep{Tibs:regr:1996,BuhlGeer11}
and compressed sensing (noiseless setting) \citep{Donoho:CS:06,Candes:Romberg:2006} literatures  that some thresholding estimators have
the ability to identify with high probability all significant
covariates from a very large set, and are numerically tractable thanks to convexity.
For the noisy situation (\ref{eq:linearmodel}) we are interested in, the lasso estimate is defined by
\begin{equation} \label{def:lasso}
\hat {\boldsymbol \beta}(\lambda) = {\rm argmin}_{{\boldsymbol \beta} \in {\mathbb R}^P} \frac{1}{2} \| {\bf y} - X {\boldsymbol \beta} \|_2^2 + \lambda \| {\boldsymbol \beta} \|_1
\end{equation}
for a positive $\lambda$.
Key assumptions must be made on the  matrix $X$ and the amount of regularization $\lambda$ for the lasso to have good estimation or selection properties.
To achieve optimal rate of convergence, compatibility condition of order $s^0$ for the matrix $X$ \citep{Van:dete:2007} \citep[Theorem 6.1]{BuhlGeer11}
is sufficient.
To achieve the \emph{variable screening} property of finding
\begin{equation} \label{eq:VS}
 \hat {\cal S}_\lambda=\{p\in\{1,\ldots,P\}:\ \hat \beta_p(\lambda) \neq 0\}  \supseteq {\cal S}^0=\{p\in\{1,\ldots,P\}:\ |\beta_p^0| \geq C\}
 \end{equation}
for some relevance level $C$ \citep{BuhlGeer11},
 the stronger restricted eigenvalue condition of order $s^0$ \citep{Bi:Ri:Tsy:2009} is sufficient.
 To achieve these properties,  the conservative penalty parameter $\lambda=4\sigma \sqrt{t^2+2 \log P}$ with $t$ large is sufficient.
 Conservative bounding inequalities lead to these results. In particular, the following set plays a key role
 \begin{equation} \label{eq:calT}
   {\cal T}=\{\|2 X^{\rm T} {\boldsymbol \epsilon} \|_\infty\leq \lambda\},
 \end{equation}
 where ${\boldsymbol \epsilon}$ is the noise in (\ref{eq:linearmodel}). If $\lambda$ guarantees that $\mathbb{P }({\cal T})$ tends to one
 (along with mild assumptions),
 then $\ell_1$- and $\ell_2$-convergence and the screening property are guaranteed.
 
 These results are asymptotic and $\lambda$ is suitable in range of order $\lambda \asymp \sqrt{\log p}$.
 Our proposal is to find a non-asymptotic $\lambda$ in this suitable range, not too large to have high true positive rate, and not too small
 to have low false discovery rate (\ref{eq:TPRandFDR}).

\subsection{Selection of $\lambda$ and estimation of $\sigma$} \label{subsct:lambdasigma}


To select the regularization parameter $\lambda$, the most common methods implemented in softwares are crossvalidation \citep{Stone74CV} and
Stein unbiased risk estimation \citep{Stein:1981} for lasso 
\begin{eqnarray*}
 {\rm SURE}(\lambda)&=&\|{\bf y}- X \hat {\boldsymbol \beta}(\lambda) \|_2^2+ 2 \sigma^2 {\rm rank}(X_{{\cal E}_\lambda}),
 \end{eqnarray*}
where ${\cal E}_\lambda$ is the equicorrelation set of $X \hat {\boldsymbol \beta}(\lambda)$ \citep{Tibs:Tayl:degr:2012}, previously derived
when ${\rm rank}(X)=P$ \citep{ZHT07}.
Being based on the $\ell_2$-loss between ${\bf y}$ and $X \hat {\boldsymbol \beta} $, they are both concerned with
 criterion~1 discussed above: prediction.
The same is true with subset variable selection for which the size $\lambda:=k={\rm card}(\{\hat \beta_{p}\neq 0, \ p=1,\ldots,P\})$ of the model is chosen
to minimize the AIC or BIC criteria \citep{AkaikeAIC73,Schw:esti:1978}:
 \begin{eqnarray*}
 {\rm AIC}(k,\sigma)&=&-2 \log {\cal L}(\hat {\boldsymbol \beta}(k), \sigma; {\bf y})+ 2k , \\
 {\rm BIC}(k,\sigma)&=&-2 \log {\cal L}(\hat {\boldsymbol \beta}(k), \sigma; {\bf y})+ k \log N,
 \end{eqnarray*}
 where $\log {\cal L}(\hat {\boldsymbol \beta}(k), \sigma; {\bf y}) = -N/2 \log \sigma^2 -N/2 \log (2\pi) -\frac{1}{2\sigma^2} \|{\bf y}-X\hat {\boldsymbol \beta}(k) \|_2^2$
 is the log-likelihood.
 BIC is known to select few covariates, and so is also geared towards the second criterion.
 A modified Monte Carlo cross validation method also aims at the second criterion  \citep{YUFENG-CVlasso14}.
 
 Since AIC corresponds to SURE when ${\rm rank}(X)=P$ and $\sigma$ is known, BIC could also potentially be considered as a criterion to select $\lambda$, although no theory supports this yet.
 Also, a practice for good prediction is to select a model with lasso for some $\lambda$, and then do least squares on the selected set or
 adaptive lasso \citep{Zou:adap:2006}; see the discussion of \citet[Section~2.8]{BuhlGeer11}. The goal of these two approaches is to shrink less the selected coefficients, so as 
 to achieve better prediction.


With the exception of cross validation, these selection rules for $\lambda$ require knowledge of the variance $\sigma^2$ of the noise.
The standard formula to estimate the variance is
\begin{equation} \label{eq:sigma2hat}
 \hat \sigma^2
 =\frac{1}{N-k} \| {\bf y} - X  \hat {\boldsymbol \beta}(\lambda) \|_2^2.
\end{equation}
Most past and current estimates are based on this formula, where $\hat {\boldsymbol \beta}(\lambda)$ and $k$ remain to be determined.
It is well known that when $\hat {\boldsymbol \beta}(\lambda)= \hat {\boldsymbol \beta}^{\rm MLE}$ solves the ``normal equation'' $X\T X  {\boldsymbol \beta} = X\T {\bf y}$,
then choosing $k=0$ amounts to the MLE of $\sigma^2$ and choosing $k={\rm rank}(X)$ provides an unbiased estimate of $\sigma^2$.
The use of this formula requires a word of caution though.
First we need ${\rm rank}(X)<N$, otherwise the estimate of variance is zero for $k=0$ and undetermined for $k=N$.
This condition is rarely satisfied in most applications when $P>N$.
Second assuming ${\rm rank}(X)<N$ and choosing $k={\rm rank}(X)$ in~(\ref{eq:sigma2hat}),
one can easily show that the estimate of $\sigma^2$ is unbiased, using the property that $\hat {\boldsymbol \beta}^{\rm MLE}$
is  \emph{linear} (that is, a linear function of the response). But this property no longer holds if the estimator  $\hat {\boldsymbol \beta}(\lambda)$ is not linear.
It is important to recognize that best subset variable selection, the Dantzig selector and lasso are nonlinear estimators, and that
using (\ref{eq:sigma2hat}) with these estimators is just a proxy, which is known to sometimes fail.
Indeed, nonlinear estimators tend to generate various competing biases:
\begin{itemize}
 \item an upward bias if the selected model does not include the true model~(\ref{eq:Model*});
 \item an upward bias even if the selected model is exact  and if the estimate $\hat {\boldsymbol \beta}(\lambda)$ biases by shrinking towards zero (e.g., lasso).
 In that case the residuals are larger than expected and lead to overestimation of the variance;
  \item a downward bias if  unnecessary covariates collinear with ${\bf y}$ have been wrongly added into the true model.

 \end{itemize}
Nevertheless formula~(\ref{eq:sigma2hat}) is used for lasso or subset selection, with more or less success.
For subset selection when $P>N$, optimizing AIC or BIC over both $k$ and $\sigma$ gives poor results, due to the bad combination of formula \eqref{eq:sigma2hat}
with a nonlinear estimate.
For lasso, \citet{ReidTibshFriedarchiv14} discuss and empirically study the case where $\lambda$ is selected by cross validation and $k$ is the number of non-zero coefficients in the corresponding lasso estimate, and
conclude that this estimate of variance gives good results on their simulations but lacks theoretical grounds.
They also consider scaled lasso \citep{scaledlasso12} that jointly minimizes
$$
{\cal L}_{\lambda_0,a}({\boldsymbol \beta},\sigma)=\frac{1}{2N\sigma^2} \| {\bf y} - X {\boldsymbol \beta}\|_2^2+\frac{(1-a)\sigma}{2}+\lambda_0 \| {\boldsymbol \beta} \|_1
$$
for $\lambda_0=\sigma \sqrt{2^{j-1} \log P}$ with some arbitrary $j\in \{1,2,3\}$ and $a$,
by alternating between updating ${\boldsymbol \beta}$ and $\sigma$. At each iteration the solution in $\sigma^2$ is formula~(\ref{eq:sigma2hat}) with $\lambda=\hat \sigma \lambda_0$ and $k=aN$.
Under certain conditions they prove consistency and asymptotic normality of their variance estimator.
 \citet{Fan:Guo:Hao:2012} propose refitted cross validation (RCV), also based on lasso and formula~(\ref{eq:sigma2hat}).  Their idea is to select two models on two cross validated data sets
 and to use the estimated models ${\cal M}_1$ and ${\cal M}_2$ on data sets 2 and 1, respectively, to get two variance estimates and average them.
 They prove asymptotic normality under some conditions.
 
%
%
 
 \subsection{Our contribution}
   
 The contribution of this paper is a new way to select the threshold $\lambda$ for thresholding methods like lasso or the Dantzig selector.
 It is not based on resampling like cross validation, and contrarily to most existing methods,
 it does not require calculation of $\hat {\boldsymbol \beta}(\lambda)$ for many $\lambda$'s until a optimal criterion is achieved.
 Instead $\hat {\boldsymbol \beta}(\lambda)$ must be calculated for a single $\lambda$, which makes our method also appealing from a computational point of view.
 The new selection rule aims at selecting a good set of covariates of a size not too small to have high true positive rate and a good predictive performance,
 and not too large to have low false discovery rate.
 It is based on the ideas of \emph{universal threshold} for wavelet smoothing \citep{Dono94b} and localization of the \emph{bulk edge} for low rank matrix estimation \citep{Donoho:SVDHT:2013}.
 We define this new method in Section~\ref{subsct:QUT} and study its variable selection performance in Section~\ref{subsct:LOI}
 with a massive Monte Carlo simulation inspired by results in compressed sensing; in particular we observe a phase transition in the probability for lasso
 to include the right model for some oracle threshold $\lambda$, and that the proposed selection rule for $\lambda$ achieves a comparable phase transition.
 In Section~\ref{sct:numerical}, we investigate  on two real data sets and on the simulation of \citet{ReidTibshFriedarchiv14} how our selection of $\lambda$ compares
 with existing selection rules.
 We also apply it to a nonparametric wavelet-based estimator to solve an inverse problem involving the Abel operator, for which 
 lasso is employed to impose sparsity on the wavelet coefficients.
 We then draw some conclusions in Section~\ref{sct:conclusion}.

\section{Quantile universal threshold}
\label{sct:MSwithQUT}

\subsection{Definition}
\label{subsct:QUT}

Given responses ${\bf y}$ and a regression matrix $X$, many thresholding methods share the property that there exists a finite threshold $\lambda$
for which the corresponding estimate is fully sparse,
that is, when all estimated coefficients  are equal to zero: $\hat {\boldsymbol \beta}(\lambda)={\bf 0}$.
For lasso~(\ref{def:lasso}), the Dantzig selector \citep{Cand:Tao:dant:2007} and Waveshrink \citep{Dono94b},
this \emph{thresholding statistic} is $\lambda=\|X\T {\bf y} \|_\infty$.
This is also true for group lasso \citep{Yuan:Lin:mode:2006} with another statistic.
The thresholding statistic can be inferred from the dual problem, for instance derived by \citet{Osbo:Pres:Turl:on:2000} for lasso.

Based on this property,
\citet{Dono94b} proposed the \emph{universal threshold} $\lambda_N=\sigma \sqrt{2 \log N}$
for their Waveshrink estimator, a particular case of lasso where $X$ is an $N\times N$ orthonormal matrix.
With that choice of a threshold, Waveshrink sets all coefficients to zero with high probability when the data indeed come from the null model.
More precisely one can show that
\begin{equation} \label{eq:propH0}
\mathbb{P}(\hat {\boldsymbol \beta}({\lambda_N})={\bf 0} \mid {\boldsymbol \beta}^0={\bf 0})
=\mathbb{P}(\|X^{\rm T}{\bf Y}\|_\infty \leq \lambda_N )\stackrel{\cdot}=1-\alpha_N,
\end{equation}
with $\alpha_N=1/\sqrt{\pi \log N}$.
In other words, the universal threshold $\lambda_N$ is the $(1-\alpha_N)$-quantile of the distribution of
$\Lambda=\|X^{\rm T}{\bf Y}\|_\infty$ under the null hypothesis
that ${\bf Y} \sim {\rm N}({\bf 0},\sigma^2 I_N)$.
Note that the thresholding statistics $\Lambda=\|X^{\rm T}{\bf Y}\|_\infty$ is intimately connected to the set ${\cal T}$ defined in (\ref{eq:calT}),
since when $ {\boldsymbol \beta}^0={\bf 0}$ then ${\bf Y}={\boldsymbol \epsilon}$.
Such a threshold $\lambda_N$ also has good power: for instance under the alternative that only the first entry $\beta_1^0= \tau \sigma$ with $\tau>0$,
then $\mathbb{P}(\hat { \beta}_1(\lambda_N) \neq 0)\stackrel{\cdot}= 1-\Phi(\sqrt{2 \log N}-\tau)$ 
grows to one quite fast with $\tau$.
In fact Waveshrink satisfies an oracle inequality \citep{Dono94b} and has good minimax properties \citep{Dono95asym}. 

The good properties of the universal threshold motivate its extension to the situations when the matrix $X$ is not necessarily orthonormal, in particular when $P>N$.
We call this extension the \emph{quantile universal threshold} that we now define.
%

\bigskip

{\bf Definition}. Let $X$ be any $N \times  P$ matrix of covariates. The \emph{quantile universal threshold} $\lambda_{{\rm QUT}}$ is defined as
\begin{equation} \label{eq:defQUT}
\lambda_{{\rm QUT}}=\sigma F_{\Lambda}^{-1}(1-\alpha_P)
\end{equation}
with $\alpha_P=1/\sqrt{\pi \log P}$, where $F_{\Lambda}$ is the c.d.f.~of $\Lambda({\bf Y})=\|X^{\rm T} {\bf Y} \|_\infty$ when data follow the null  model ${\bf Y}\sim {\rm N}({\bf 0},I_N)$.

\bigskip

The quantiles of $F_{\Lambda}$ are rarely known in closed form, except for special matrices like for Waveshrink discussed above 
and for total variation smoothing \citep{ROF92}.
For the former, the universal threshold $\lambda_N=\sigma \sqrt{2 \log N}$ controls the maximum of $N$ i.i.d.~standard Gaussian distribution, and for the latter 
the universal threshold $\lambda_N=\sigma \sqrt{N/4 \log \log N}$ controls the maximum of a discretized Brownian bridge  \citep{SardyTseng04}.
For low rank matrix estimation by hard thresholding singular values, \citet{Donoho:SVDHT:2013} derive an asymptotic threshold $\lambda_N=4\sqrt{N}/\sqrt{3}$
at the \emph{bulk edge} of the distribution of the singular values under the null hypothesis.

QUT is the finite sample counterpart of these asymptotic theoretical thresholds. QUT can be employed without requiring a specific structure for $X$. It is simply
based on the distribution of the thresholding statistic and its empirical evaluation.
Indeed for any matrix $X$, the quantile universal threshold of $F_{\Lambda}$  can be calculated by Monte Carlo:
generate ${\bf y}^{(m)}$ from ${\rm N}({\bf 0},I_N)$
for $m=1,\ldots,M$, calculate $\lambda^{(m)}=\|X^{\rm T} {\bf y}^{(m)} \|_\infty$, and take the $(1-\alpha_P)$-empirical quantile  of the $\lambda^{(m)}$.
These steps require simple calculations, lasso must be solve for a single $\lambda$, and no cross validation or bootstrap resampling technique is involved.

Finally we commented in Section~\ref{subsct:lambdasigma} that no theory has been established to use BIC with lasso.
But when $X$ is orthonormal and $\sigma$ is known, one can compare BIC with QUT,
as one can show that the following inequality holds for the penalty parameter of hard thresholding 
\begin{equation} \label{eq:QUTvsBIC}
\lambda_{{\rm QUT}}^{\ell_0}\stackrel{\cdot}{=}2 \log N> \log N=\lambda_{{\rm BIC}},
\end{equation}
where $\lambda_{{\rm QUT}}^{\ell_0}$ is QUT's penalty for best subset variable selection. This result can be inferred
by considering $\ell_p$-regularization and letting $p$ tend to zero \citep{SardySLIC09}.
This shows here that QUT is more conservative than BIC, and leads to a sparser estimate.

\subsection{Empirical evidences with $\sigma$ known}
\label{subsct:LOI}

QUT satisfies property~(\ref{eq:propH0}) under the assumption that all coefficients are null. In this section, we investigate other probability properties when a certain proportion of coefficients are non-zero.
We concentrate on the lasso estimator.

We rely on an experiment designed by \citet{preciseunder10} for compressed sensing, that we extend here to noisy data.
Their experiment is simple, but rich of interesting observations.
In a nutshell, they consider $N \times P$ i.i.d.~standard Gaussian matrices $X$ with a  fixed number of columns $P = 1600$.
Two parameters vary: the number of rows $N$ and
the number $k:=s^0$ of non-zero entries in ${\boldsymbol \beta}^0$.
Here $k$ plays the role of $s^0$, the cardinality of ${\cal S}^0$ defined in (\ref{eq:Model*}).
For every pair $(N,k)$, they simulate a large number of matrices $X$, and based on ${\bf y}=X {\boldsymbol \beta}^0$, they estimate the probability of retrieving ${\boldsymbol \beta}^0$ by Monte Carlo.
Then they plot those probabilities as a function of
\begin{eqnarray} \label{eq:deltarho}
 \delta=N/P:&& \mbox{the undersampling factor}, \\
 \rho=k/N: && \mbox{the sparsity factor},
\end{eqnarray}
for $k \in \{1,\ldots,N\}$ and $N$ taking values on a grid ranging from $N=160$ up to $N=1440$ in nine equal steps.
What they observe are two regions and a phase transition between them:  the probability is equal to one in a region where $\rho$ is small and $\delta$ is large, and abruptly drops to zero in the other region.
This means that when the amount of information $N$ is small compared to the number of unknowns $P$, or when the number of nonzero parameters $k$ is large compared to $N$,
then retrieving the vector ${\boldsymbol \beta}^0$
is impossible with a procedure that is essentially lasso for noiseless data.
In the other region, the true model is found with probability one.

\begin{figure}[!ht]
   \begin{center}
   \includegraphics[height=7cm, width=13cm]{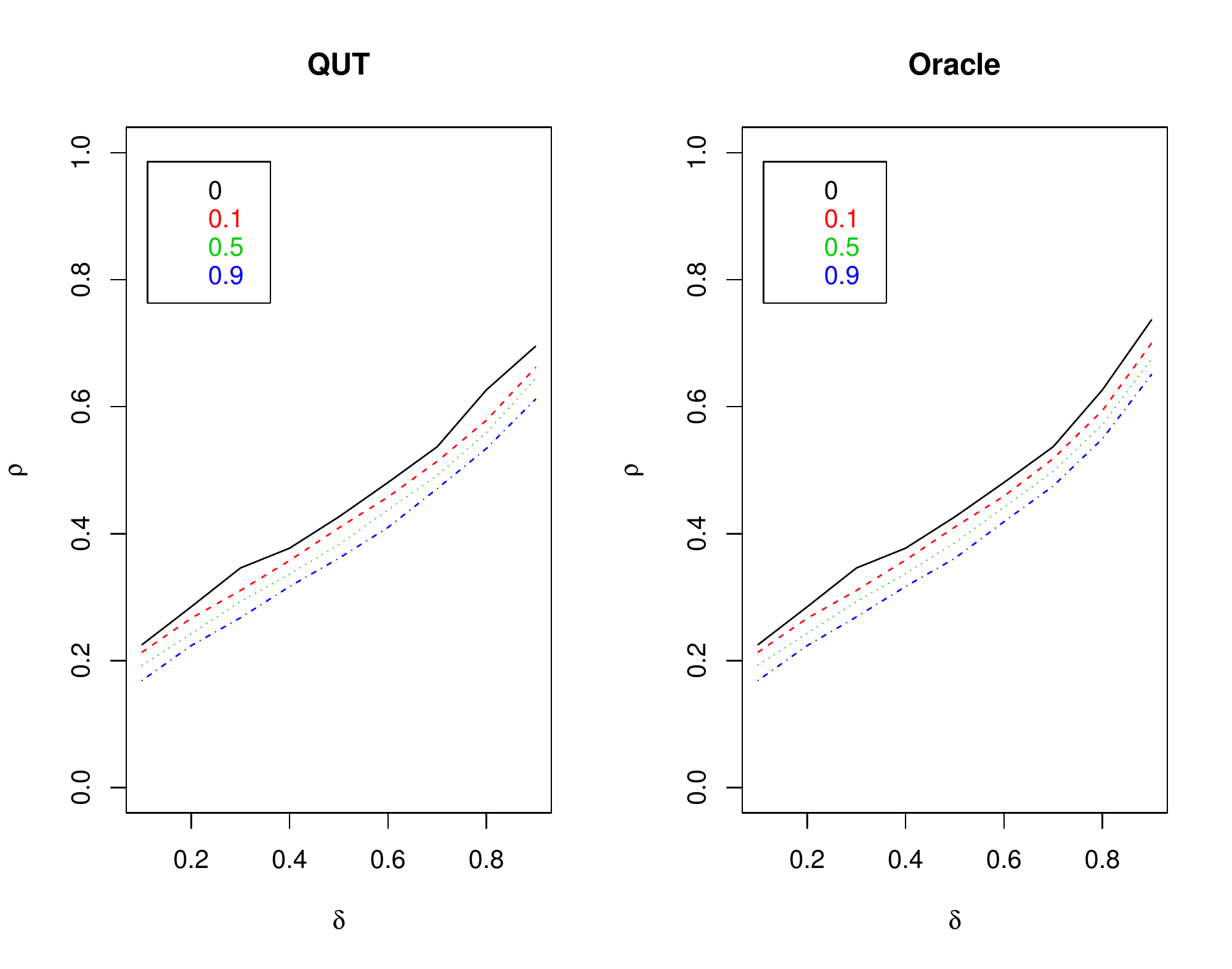}
   \caption{Phase Transition in the oracle inclusive probability (OIP) as a function of $\delta=N/P$ and $\rho=k/N$.
   Left: with the quantile universal threshold (QUT). Right: with the oracle inclusive threshold.
   Level sets of probabilities of being Oracle Inclusive of 90\%, 50\%, 10\%, and 0\% are indicated in blue, green, red and black curves, respectively.
   \label{fig:phaseTransition}}
   \end{center}
 \end{figure}
 
We perform the same simulation but with standard Gaussian noise according to model~(\ref{eq:linearmodel}).
The value of the $k$ non-zero coefficients is set to the large value of ten times the standard deviation of the noise to make them highly identifiable by lasso,
when possible.
For each instance of $(N,k)$ we repeat $100$ times the experiment, resulting in a massive simulation study calculating more than $300'000$ lasso estimates.
To parallel their study, we consider two characteristics of interest in regression:
\begin{itemize}
 \item the oracle inclusion property: we say that an estimator with estimated coefficients $\hat {\boldsymbol \beta}(\lambda)$ is \emph{oracle inclusive} if
  \begin{equation} \label{eq:OI}
 \hat {\cal S}_\lambda=\{p\in\{1,\ldots,P\}:\ \hat \beta_p(\lambda) \neq 0\}  \supseteq {\cal S}^0=\{p\in\{1,\ldots,P\}:\ \beta_p^0 \neq 0\}.
 \end{equation}
 A thresholding estimator may not be oracle inclusive
 either because the chosen model is too small, or there is no model in the class of models selectable by the estimator that includes the correct model.
 The oracle inclusive property corresponds to the \emph{variable screening} property (\ref{eq:VS}) of relevance level $C=0$.
 \item the oracle inclusive ratio. Suppose that, for a given sample, there exists a lasso model ${\cal S}_{\lambda^*}$ of cardinality $s^*$
 that is oracle in the sense that  it includes the correct model and that it is of smallest cardinality among all oracle inclusive models.
 In practice, we would like lasso to find that smallest oracle inclusive model, so as to  have high TPR and low FDR.
 Consider now a method to select $\lambda$ without oracle, for instance cross validation or QUT, and
 let ${\hat s}_\lambda$ be the cardinality of the corresponding  model, which is set to $\hat s_\lambda=+\infty$ if the estimated model is not oracle inclusive.
 The oracle inclusive ratio is defined by
 \begin{equation} \label{eq:OIR}
   {\rm OIR}=\frac{|{\cal S}_{\lambda^*}|}{|\hat {\cal S}_\lambda|}=\frac{s^*}{{\hat s}_\lambda} \leq 1.
 \end{equation}
 This ratio measures the tightness of the selected model to the smallest oracle inclusive model.
 
 {\bf Property}: a model with ${\rm OIR}=1$ has the lowest FDR among all models with ${\rm TPR}=1$.
 
 With best subset variable selection (that visits all $2^P$ models) this ratio is always one.
 With forward/backward variable selection, it is less or equal to one. 
 With lasso, the closer this ratio is to one, the less unnecessary covariates are included in the estimated oracle inclusive model.
\end{itemize}

\begin{figure}[!ht]
   \begin{center}
   \includegraphics[height=10cm]{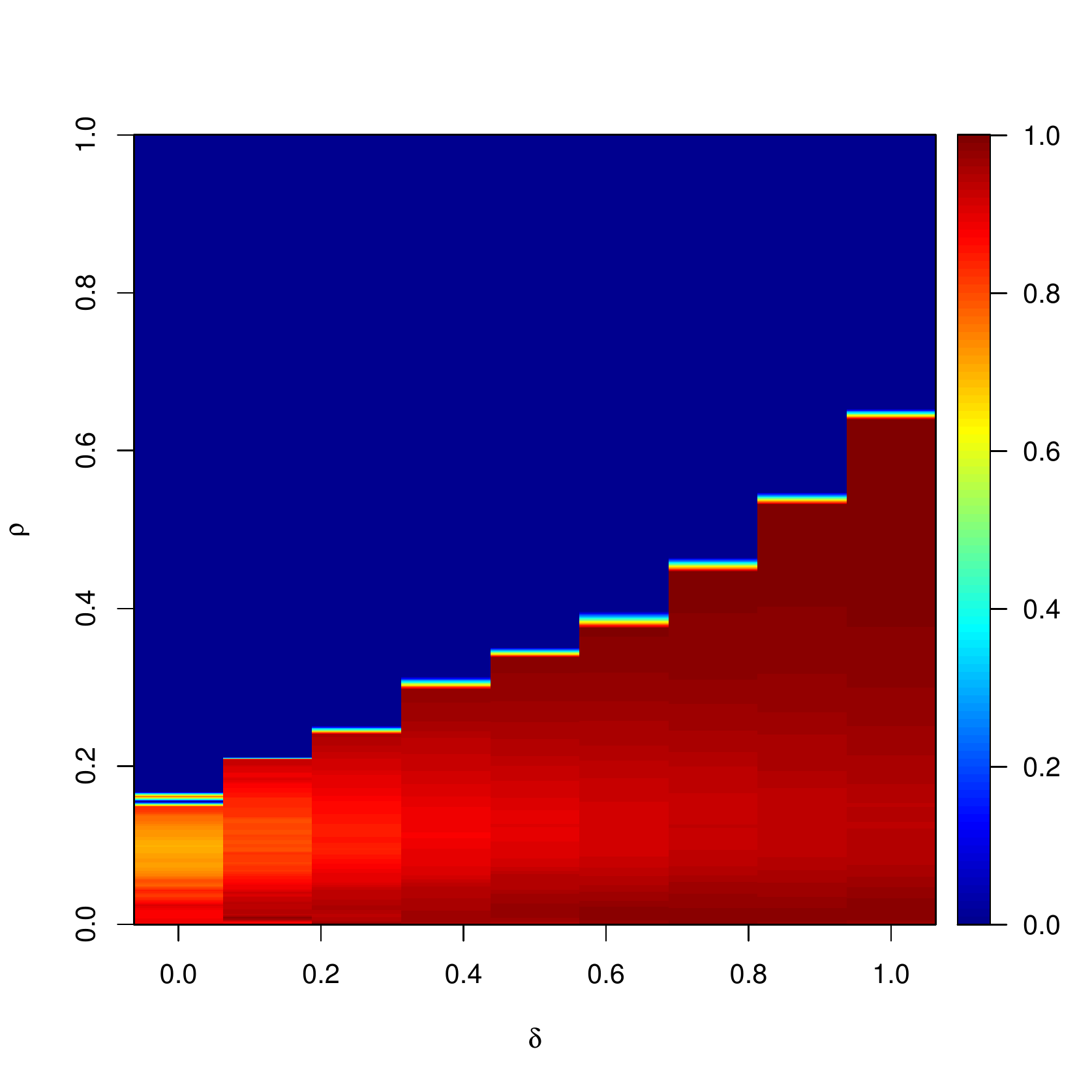}
   \caption{Phase transition for oracle inclusive ratio (OIR) for the quantile universal threshold (QUT) for lasso  as a function of $\delta=N/P$ and $\rho=k/N$. Values are shown color in based plot starting from blue (0) to red (1).
   \label{fig:LOIphase}}
   \end{center}
 \end{figure}

Considering lasso in particular and $\sigma=1$ known, Figure~\ref{fig:phaseTransition} plots the oracle inclusion probability as a function of $(\delta,\rho)$ when $\lambda$ is selected with QUT (left)
and when a search over $\lambda$ is performed to see whether any lasso model is oracle inclusive (right).
The results are very similar to the noiseless case observed by~\citet{preciseunder10} with a clear phase transition.
Looking at the right plot, we see that oracle lasso has a good oracle inclusive property, and comparing the left with the right plots, we see that QUT is almost as oracle inclusive as it is possible.
To investigate whether QUT does not include too many unnecessary variables, Figure~\ref{fig:LOIphase} plots the oracle inclusive ratio for QUT:
we see that OIR is near one in the region of high oracle inclusion probability. This means that QUT offers a good compromise between true positive and false discovery rates.

\begin{figure}[!ht]
   \begin{center}
   \includegraphics[height=9cm]{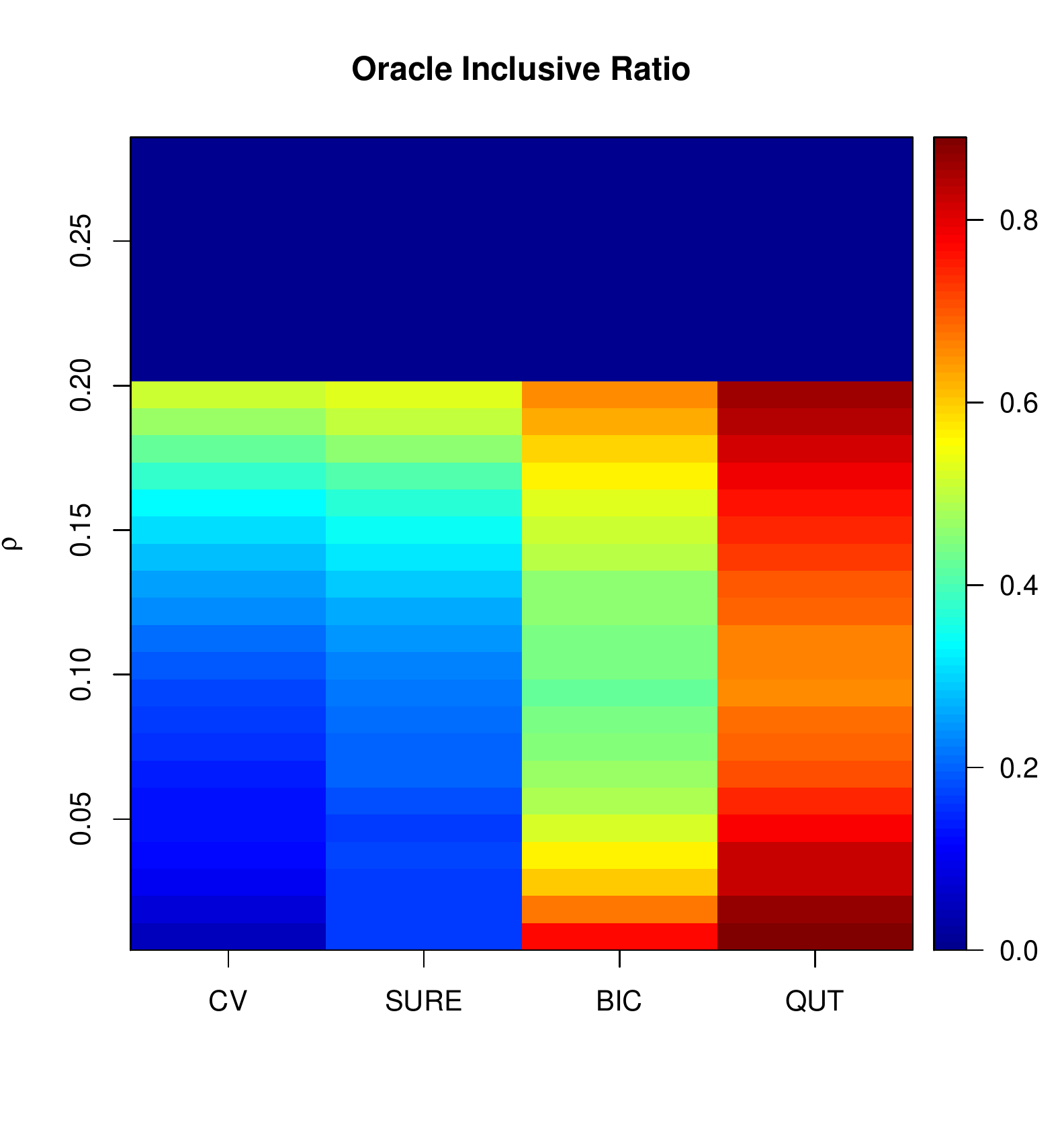}
   \caption{Comparison of the oracle inclusive ratio OIR between three selections of $\lambda$ for lasso, as of function of $\rho=k/N$. Here $\delta=N/P=0.2$ is fixed
   and corresponds to Figure~\ref{fig:LOIphase} at $\delta=0.2$.
   \label{fig:LOIdelta}}
   \end{center}
 \end{figure}

Cross validation would also have a similar phase transition diagram as Figure~\ref{fig:phaseTransition},
because it tends to choose a small $\lambda$, and hence to include the correct model.
Cross validation is far from having a good OIR however.
To see this for a specific $\delta=N/P=0.2$, Figure~\ref{fig:LOIdelta} plots  oracle inclusive ratio as a function of  $\rho=k/N$
for CV, SURE, BIC and QUT.
Cross validation and SURE have poor oracle inclusive ratio compared to BIC, and QUT is nearly optimal in that regime.

The phase transition diagram points to a weakness of cross validation because it replaces the sample size $N$ by a sub-sample of size $N'<N$.
Consequently $\delta'=N'<P$ is smaller and $\rho'=k/N'$ is larger, making it
more likely to move into a region where lasso has zero probability of being oracle inclusive.
 A possible way to cope with this problem is to perform sure independent screening \citep{Fan:Lv:sure:2008}.

\section{Data analysis}
\label{sct:numerical}

This section considers data when $\sigma$ is unknown and must be estimated to select the lasso model.
The threshold selection rules for lasso considered are: 10-fold cross validation (CV), modified Monte Carlo cross validation (MCCV) \citep{YUFENG-CVlasso14},
universal threshold (QUT),
Bayesian information criterion (BIC), Stein unbiased risk estimation (SURE), scaled lasso (SL) and stability selection (SS) \citep{stabsel10}.
We used the code provided by the authors for MCCV, SL and SS.
Cross validation and stability selection do not require estimation of $\sigma$ and serve as benchmark: the first one has good predictive risk and good true positive rate,
and the second one has good false discovery rate. This means that the number of non-zero coefficients should lie between CV and SS,
and that a method approaching the predictive risk of CV with less non-zero coefficients offers a good compromise between high TPR and low FDR.

\subsection{Real data analysis}
\label{subsct:real}

Knowing whether the relevant covariates have been correctly detected from real data is impossible.
What can be compared are the number of selected covariates and the corresponding predictive performance based on a training and test sets.
For two choices of $\lambda$ with comparable predictive performances,
the sparsest model is preferable. The benchmark for predictive performance is certainly cross validation, which tends to include too many variables however.
We consider two data sets:
\begin{itemize}
 \item {\tt Riboflavin} data set of \citet{PeterBulbiology:14} with $N=71$ measurements of production rate in Bacillus subtilis and $P=4088$
gene expressions.
 \item {\tt Chemometrics} data set of \citet{SardyISI08} with $N=434$ measurements of fuel octane level and $P=351$ spectrometer measurements.
\end{itemize}

\begin{figure}[!ht]
   \begin{center}
   \includegraphics[height=10cm]{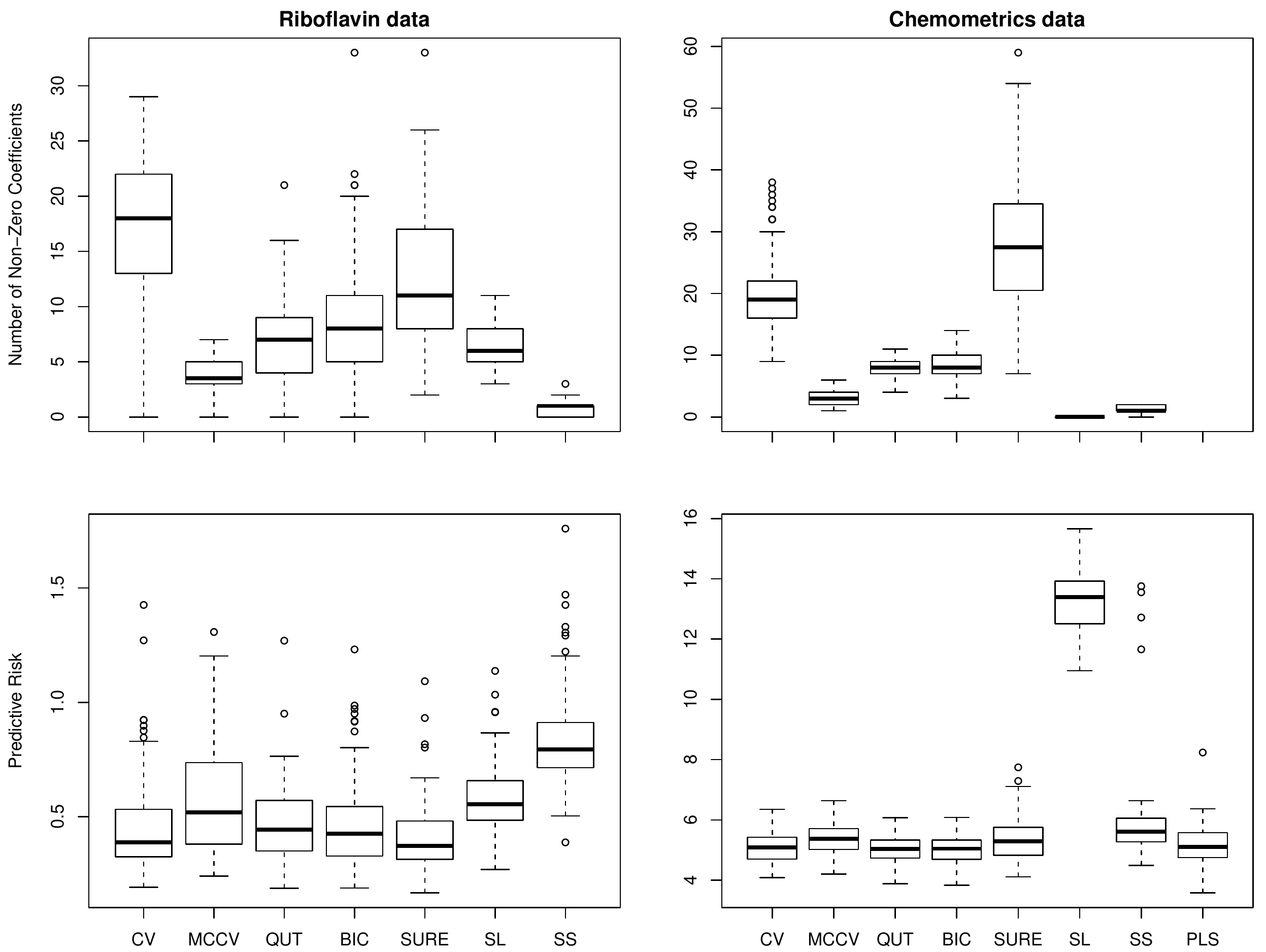}
   \caption{Based on Riboflavin (left) and Chemometrics (right) data, boxplots of one hundred results obtained by a half split of the data into training and tests sets:
   number of non-zeros coefficients on training sets (top) and predictive risk on test sets (bottom).
   \label{fig:ribochemo1}}
   \end{center}
 \end{figure}
 
For each one, we repeat one hundred times the splitting of the data set into a training and a test sets of equal sizes.
This gives undersampling factors of very low $\delta=0.87\%$ and high $\delta=62\%$, respectively.
For each training set we estimate the variance with RCV discussed in Section~\ref{subsct:lambdasigma}.
For each training and test sets, we record the number of non-zero coefficients selected
on the training set, as well as the corresponding predictive risk on the test set.
To improve the predictive risks, we fit the final model by least squares with the covariates selected by the respective methods.
The boxplots of these one hundred values represented on Figure~\ref{fig:ribochemo1}
corroborates the results of Section~\ref{subsct:LOI} in that QUT has the smallest number of non-zero coefficients
for a predictive risk comparable to CV.
MCCV tends to select smaller models than QUT, but its predictive risk is not as good, pointing to the fact that some variables must be missing.
This indicates that QUT captures the right complexity of these two data sets.
We also considered partial least squares (right boxplot for Chemometrics data), the method of choice for chemometricians, yet comparable to lasso with QUT.
Finally, we see that QUT is more conservative than BIC, which corroborates inequality~(\ref{eq:QUTvsBIC}).

\begin{figure}[!ht]
   \begin{center}
   \includegraphics[height=6cm]{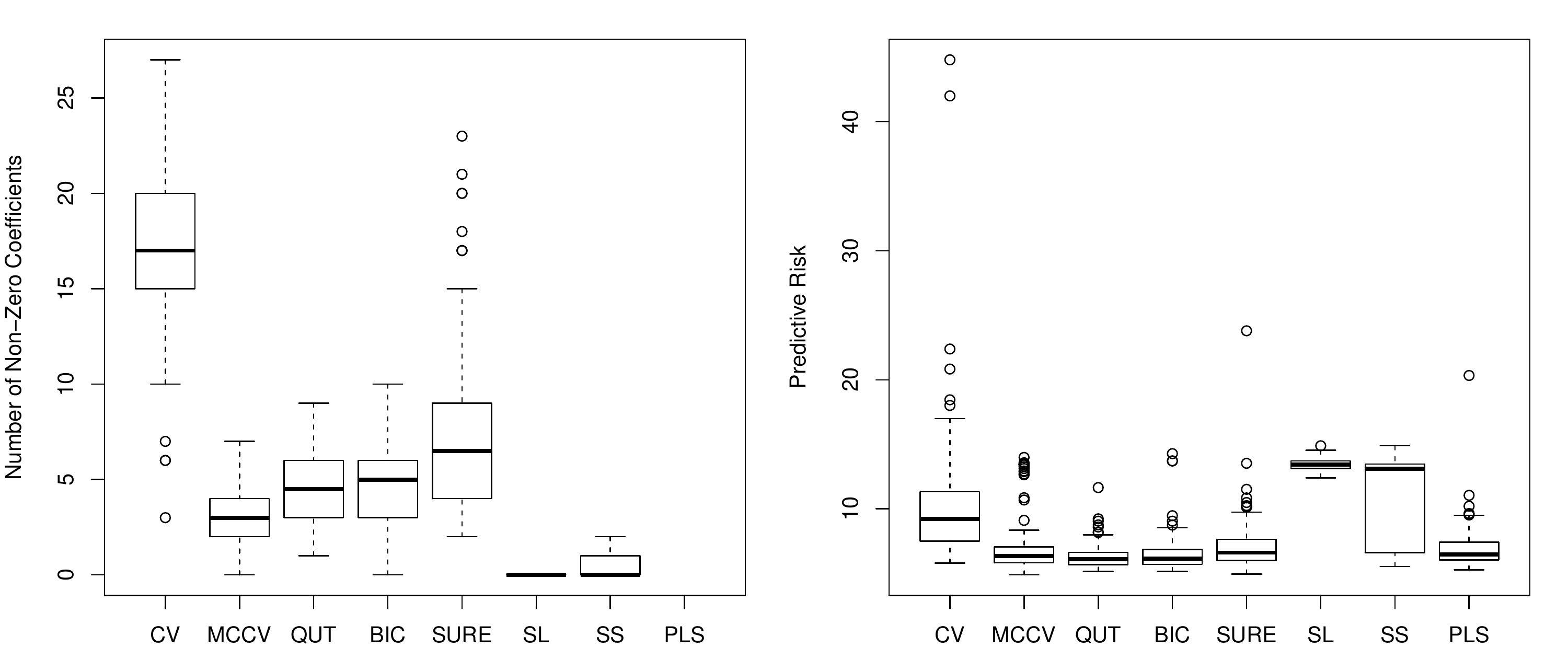}
   \caption{Simulation based on Chemometrics data: same outputs as for the right plots of Figure~\ref{fig:ribochemo1}, but with only $10\%$ training data
   as opposed to $50\%$ before.
   \label{fig:chemo2}}
   \end{center}
 \end{figure}

For the Chemometrics data, we also consider a smaller undersampling factor of about $\delta=12\%$ by taking only $10\%$ of the data for the training set and keeping $90\%$ for the test set.
This leaves 36 observations for training with 10-fold cross validation.
Figure~\ref{fig:chemo2} shows how cross validation collapses, and how QUT outperforms all other lasso models as well as partial least squares.
Note also how MCCV improves over standard CV.

\subsection{Simulation analysis}
\label{subsct:simu}

\begin{figure}[!ht]
   \begin{center}
   \includegraphics[height=9cm]{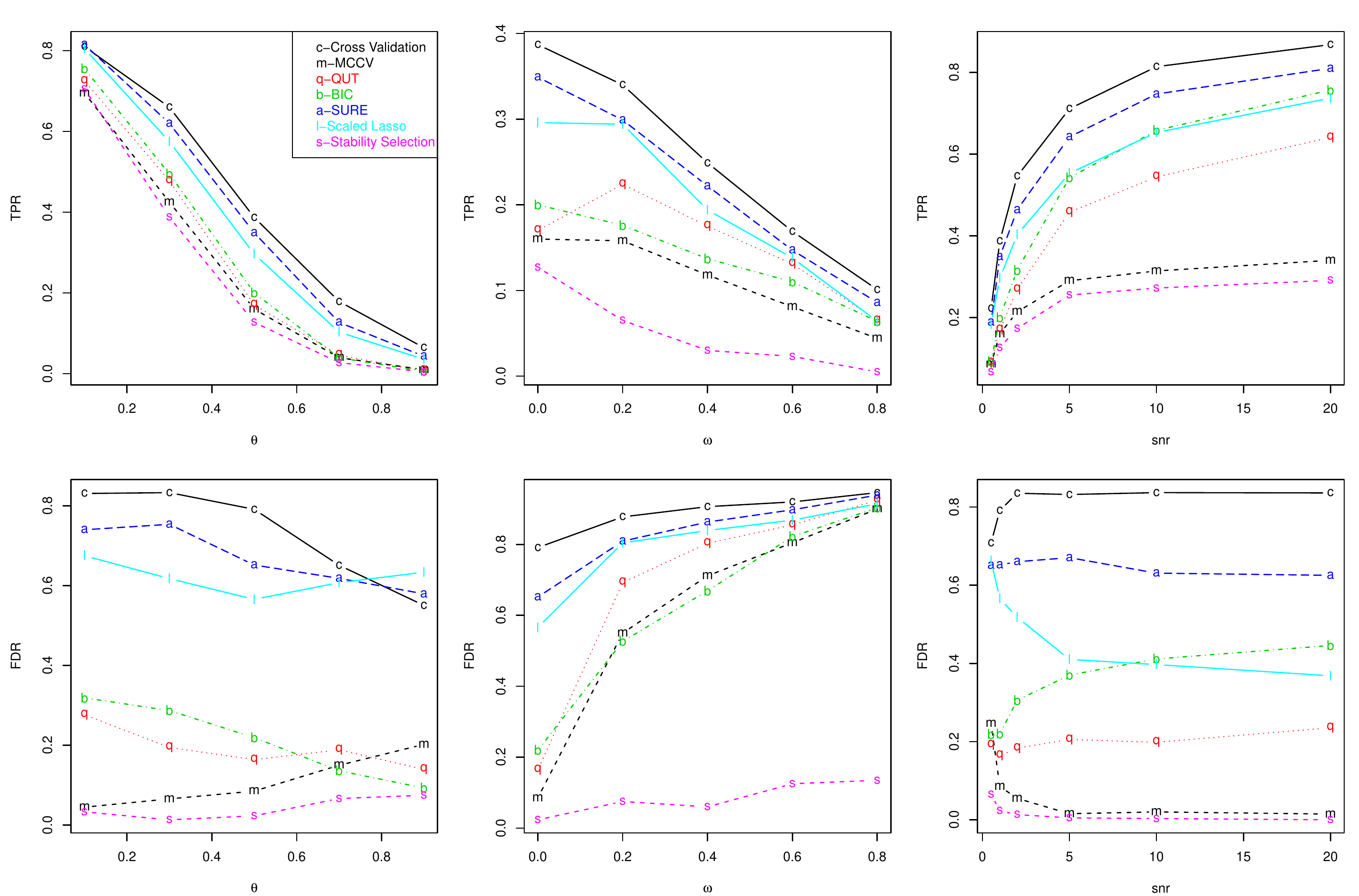}
   \caption{TPR and FDR plots showing the effect of changing sparsity parameter $\theta$, correlation  $\omega$ in the columns of $X$,
   and snr. Results show the median value for six selection rules: CV, QUT, BIC, SURE, SL and SS .
   \label{fig:tprfdr}}
   \end{center}
 \end{figure}

In order to explore further the selection rules for the threshold of lasso,
we perform simulations based on \citet{ReidTibshFriedarchiv14}.
All simulations are run under a fixed sample size of $N=100$, and fixed number of covariates $P=1000$, with unite noise variance.
Elements of the predictor matrix $X$ are generated randomly following standard normal distribution and
all the correlations between columns of $X$ are set to the same positive value $\omega$.
Such correlation matrix $\Sigma_\omega$  guarantees variable screening asymptotically \citep{BuhlGeer11}.
The number of nonzero coefficients in the true ${\boldsymbol \beta}^0$ is set to $\lceil N^{\theta} \rceil$,
where $0\leq \theta \leq 1$, having ${\boldsymbol \beta}^0$ sparser when $\theta$ is close to 0 and denser when close to 1. 
The values of the nonzero coefficients are chosen randomly from a sample of a ${\rm Laplace}(1)$ distribution.
The indices of the nonzero coefficients are also randomly selected.
The elements of the resulting coefficients ${\boldsymbol \beta}^0$ are finally scaled such that the signal to noise ratio
${\rm snr}={{\boldsymbol \beta}^0}^{\rm T}\Sigma_\omega {\boldsymbol \beta}^0/\sigma^2$ takes specific values.
Three different sets of simulations are conducted by letting one of the parameters vary at a time:
\begin{itemize}
\item The correlation parameter $\omega\in \{ 0,0.2,0.4,0.6,0.8 \}$, for fixed $\theta=0.5$ and ${\rm snr}=1$.
\item The sparsity parameter $\theta \in \{0.1,0.3,0.5,0.7,0.9\}$, for fixed ${\rm snr}=1$ and $\omega=0$.
\item The signal to noise ratio parameter ${\rm snr} \in \{ 0.5,1,2,5,10,20 \}$, for fixed $\theta=0.5$ and  $\omega=0$.
\end{itemize}
For each setting of parameters, 100 replications were performed to estimate  TPR and FDR for the selection rules discussed
in Section~\ref{subsct:real}:
CV, MCCV, QUT, BIC, SURE, SL and SS.
Figure~\ref{fig:tprfdr} shows the median value of TPR and FDR by changing the parameters as described above.
CV, SURE and SL have poor FDR, and should not be considered if FDR is a concern.
SS has the best FDR, but the lowest TPR. As expected, QUT offers the best compromise between FDR and TPR, closely followed by MCCV and BIC.

\subsection{QUT for linear inverse problems}

\citet{DonohoWV95} proposed wavelet-based methods for solving linear inverse problems. Owing to sparse wavelet representation, least squares can be regularized by $\ell_1$-penalizing the corresponding wavelet coefficients.
Our goal is to show how QUT can also be employed in this nonparametric context
to solve ill-posed inverse problems with lasso.

 \begin{figure}[!ht]
   \begin{center}
  \includegraphics[width=\textwidth]{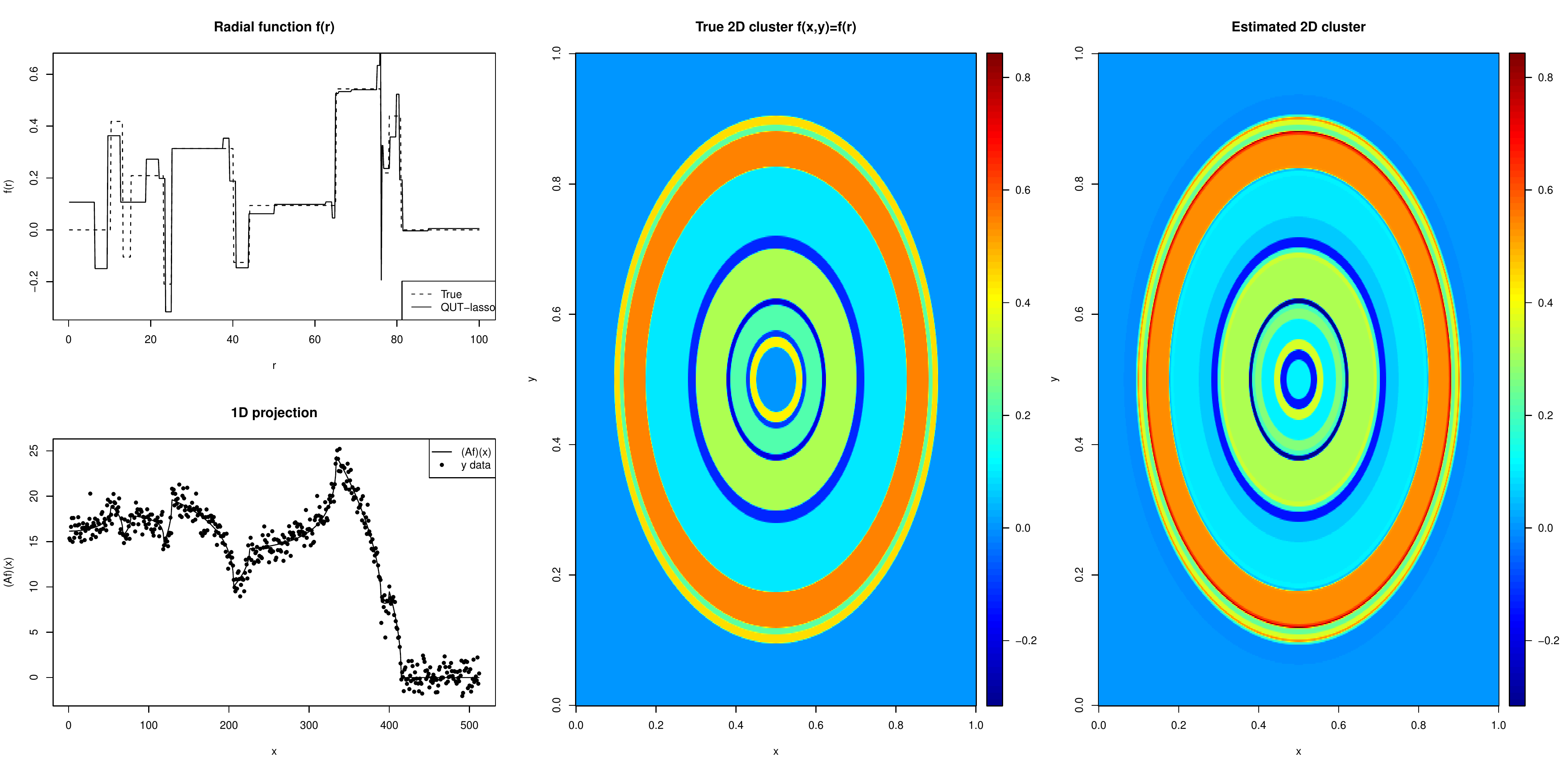}
  \caption{Illustration of one Abel simulation with ${\rm snr}=0.25$. Top-left: Radial {\tt blocks} profile function with its corresponding estimate using QUT-lasso.
   Bottom-left: 1D projection of the 2D cluster for $N=512$ and unit variance.  Center: true 2D cluster. Right: 2D cluster using radial profile estimated by QUT.
   \label{fig:abelexample}}
  \end{center}
 \end{figure}

We consider in particular the Abel problem that we encountered in a Cosmology application where massive galaxy clusters emit X-rays measured on a telescope.
The emission has an unknown 3D intensity function $f(x,y,z)$ and the telescope can be seen as a 2D set of captors facing lines until the end of cosmos.
Recovering the 3D intensities from the 2D measurements along these lines can be achieved with the assumption that the intensity function
is radial with an unknown 1D profile $f(r)$ tending to zero at infinity, hence assuming some symmetry of the galaxy clusters \citep{konrad2013}. 
We illustrae with the Abel projection from 2D to 1D for simplification: 
\begin{equation} \label{eq:abel}
(Af)(x)=2\int_{x}^\infty \frac{f(r)r}{\sqrt{r^2-x^2}}\,\mathrm{d}r
\end{equation}
models the relation between the profile function $f$ at a distance $x$ from the center of the telescope to the mean of the observed measurements $(Af)(x)$. Further assuming an expansion
of $f$ on a wavelet basis $W$, the X-ray telescope measurements ${\bf y}$ follows the linear model (\ref{eq:linearmodel}), namely:
$$
{\bf y} = X {\boldsymbol \beta}^0 + {\boldsymbol \epsilon} \quad {\rm with} \quad X=A W.
$$
For an illustration, Figure~\ref{fig:abelexample} shows the radial 2D function $f(x,y)$ in the center, the true profile function $f(r)$ (here the {\tt blocks} function, top left),
and the data ${\bf y}$ in the bottom left plot. The goal is to recover the radial function ${\bf f}=W {\boldsymbol \beta}^0$.

We employ lasso (\ref{def:lasso}) for the estimation of the sparse wavelet coefficients ${\boldsymbol \beta}^0$ after rescaling the columns of $X$ to unit variance.
For the sake of identifying true non-zero coefficients, we consider
the {\tt blocks} function  for $r$ going between 0 and 100 on a grid of $N=512$ points \citep{Dono94b} along with $P=N$ Haar wavelets (so the undersampling factor is $\delta=1$), leading to a total of 54 non-zero wavelet coefficients
(so the sparsity factor is $\rho=54/512=0.11$).
The variance is one and assumed known, and the signal to noise ratio varies in ${\rm snr} \in  \{0.25,0.50,1.0\}$.

For the selection of the regularization parameter $\lambda$, we consider QUT, BIC and SURE. For this $X$-fixed situation, we did not consider resampling methods
to select~$\lambda$.
Based on 100 replications for each snr, we estimate TPR, FDR and MSE of $\hat{\bf f}$ with respect to ${\bf f}$.
For better MSE, we refit by least squares the model selected by lasso with the $\lambda$ found either with QUT, BIC or SURE.

\begin{table*}\centering
\ra{1.10}
\small
\caption{\small Mean results of 100 replications of galaxy clusters simulation estimating the radial {\tt blocks} function with QUT, BIC and SURE
under different values of snr.}
\begin{tabular}{@{}lrrrrrrrrrrr@{}}\toprule
\cmidrule{2-12}
& \multicolumn{3}{c}{${\rm snr}=0.25$} & \phantom{abc} & \multicolumn{3}{c}{${\rm snr}=0.50$} & \phantom{abc} & \multicolumn{3}{c}{${\rm snr}=1.0$}\\
\cmidrule{2-4} \cmidrule{6-8} \cmidrule{10-12} 
Method & FDR & TPR & MSE && FDR & TPR & MSE && FDR & TPR & MSE \\ \midrule
QUT & 0.11 & 0.41 & 3.31 && 0.11 & 0.64 & 6.12 && 0.11 & 0.87 & 8.49\\
BIC & 0.21 & 0.50 & 4.50 && 0.27 & 0.77 & 7.87 && 0.29 & 0.94 & 10.19\\
SURE & 0.66 & 0.77 & 4.15 && 0.67 & 0.88 & 8.25 && 0.65 & 0.96 & 16.49\\
\bottomrule
\end{tabular}
\label{tab:abelresults}
\end{table*}


As expected, Table~\ref{tab:abelresults} shows that QUT has a better performance in terms of FDR. This corroborates the fact (\ref{eq:QUTvsBIC}) that QUT is more conservative than BIC.
Although its TPR is not the best, we see that QUT has the best MSE in all snr, which also leads us to the conclusion that QUT offers a very good compromise
between FDR and TPR, which carries to good MSE.
QUT is widely applicable in the field of inverse problems.



\section{Conclusion}
\label{sct:conclusion}

We proposed a new selection of the threshold $\lambda$ for convex thresholding methods, for instance lasso and the Dantzig selector.
The new selection seeks a threshold at the detection limit by controlling the maximal behavior of the thresholding statistics, that for lasso is $\Lambda=\|X^{\rm T}{\bf y}\|_\infty$, under the null hypothesis
that all coefficients are null. Our approach recovers theoretical bounds like the universal threshold $\sqrt{2 \log N}$ when $X$ is orthonormal,
and $\sqrt{N/4 \log \log N}$ when $X$ stems from a total variation smoother. For a general $X$ matrix, we rely on a simple Monte Carlo simulation to estimate the distribution
of the thresholding statistics under the null.

Because it is at the detection limit, the quantile universal threshold QUT provides lasso with a good compromise between high TPR and low FDR, which leads to good MSE properties.
The advantage is also computational since lasso needs to be solved only for a single $\lambda$ and no expensive resampling technique
like cross validation or boostrap is required.
More theory is needed to support our empirical findings.
We are also working on the extension of QUT to generalized linear models and to an application in Cosmology to handle deblurring as well as point source detection.

\section{Acknowledgements}

We thank Caroline Giacobino for interesting discussions. The authors are supported by the Swiss National Science Foundation.

%
%
%
%
%

\bibliographystyle{plainnat}
\bibliography{article}


\end{document}